\pdfoutput=1

\documentclass{article}

\usepackage[dblblindworkshop,final]{{neurips_2026}}

\usepackage[utf8]{inputenc}
\usepackage[T1]{fontenc}
\usepackage{booktabs}
\usepackage{threeparttable}
\usepackage{siunitx}
\usepackage{graphicx}
\usepackage[table]{xcolor}
\usepackage{microtype}
\usepackage{hyperref}
\usepackage{url}

\definecolor{mybest}{HTML}{C3a9f3}
\newcommand{\method}{STAGE}
\newcommand{\bench}{STAGE-Eval}
\workshoptitle{{SLMs for Agentic Systems, Paris, France, 2026}}

\definecolor{ClaudeViolet}{HTML}{6A1B9A}

\usepackage{subcaption}
\usepackage{wrapfig}
\newcommand{\meanstd}[2]{%
  #1{\scriptsize #2}%
}
\newcommand{\bestmeanstd}[2]{%
  \underline{#1}{\scriptsize #2}%
}

\title{STAGE: Source-grounded Text-to-JSON 
Artifact GEneration}

\author{
  Sunghee Ahn \\ Seoul National University \\ \texttt{snusnowhite@snu.ac.kr} \And
  Guijin Son \\ Seoul National University \\  \And
  Youngjae Yu \\ Seoul National University \\ 
}

\begin{document}
\maketitle

\begin{abstract}
From financial filings to clinical records, legacy industries rely heavily on long, unstructured documents to store high-value information. Reliably extracting this information into structured, machine-readable representations is a key prerequisite to making the contents accessible to automated systems. 
Yet, constructing reliable and scalable text-to-JSON training data remains challenging.
To address this gap, we propose \method{} (\textbf{S}ource-grounded \textbf{T}ext-to-JSON \textbf{A}rtifact \textbf{GE}neration), a source-grounded data generation pipeline that constructs reports and JSON/schema by using LLMs for scalable synthesis while validating ground-truth values against the underlying spreadsheet. Evaluations on \bench{}, our source-grounded benchmark with an 851-example test set, show that \method{} produces stronger training data than existing approaches. This improves Qwen3-4B exact match rate from 38.03\% to 74.27\% and value accuracy from 68.83\% to 90.69\%.\footnote{\bench{} and its evaluation code are available at \url{https://github.com/boradorish/STAGE-Eval}.}
\end{abstract}

\section{Introduction}

Unstructured documents (e.g., 10-K reports, radiology reports) have long served as primary media for storing rich domain-relevant information.
With LLMs and agentic systems emerging as powerful tools to process information~\citep{zhao2023survey}, attempts to transform information-dense unstructured documents into machine-readable structured data~\citep{lu2022unified} have accelerated, especially in JSON format~\citep{geng2025jsonschemabench, lu2025learning}.
The \textit{text-to-JSON} problem, which aims to transform legacy documents into structured JSON, was first attempted through direct prompting. This was followed by methods like constrained decoding~\citep{scholak2021picard, willard2023efficient, dong2025xgrammar} and structured-output APIs~\citep{qin2024toolllm, patil2024gorilla}, aiming for strict schema compliance. Recent works go beyond simply generating a format-compliant output by evaluating value-level fidelity~\citep{wang2025slot, zhou2025deepjsoneval, zhang2026extractbench, singh2026structured}. However, these efforts typically rely either on costly human annotation to produce gold-standard JSON outputs or on LLM-generated supervision to scale annotation. Critically, the latter introduces a circular dependency, as training an effective text-to-JSON model requires access to a model that is already capable of the same task. These limitations motivate the need for a new scalable and reliable data generation methodology.

\begin{wrapfigure}[17]{r}{0.48\linewidth}
    \vspace{-0.6\baselineskip}
    \centering
    \includegraphics[width=\linewidth]{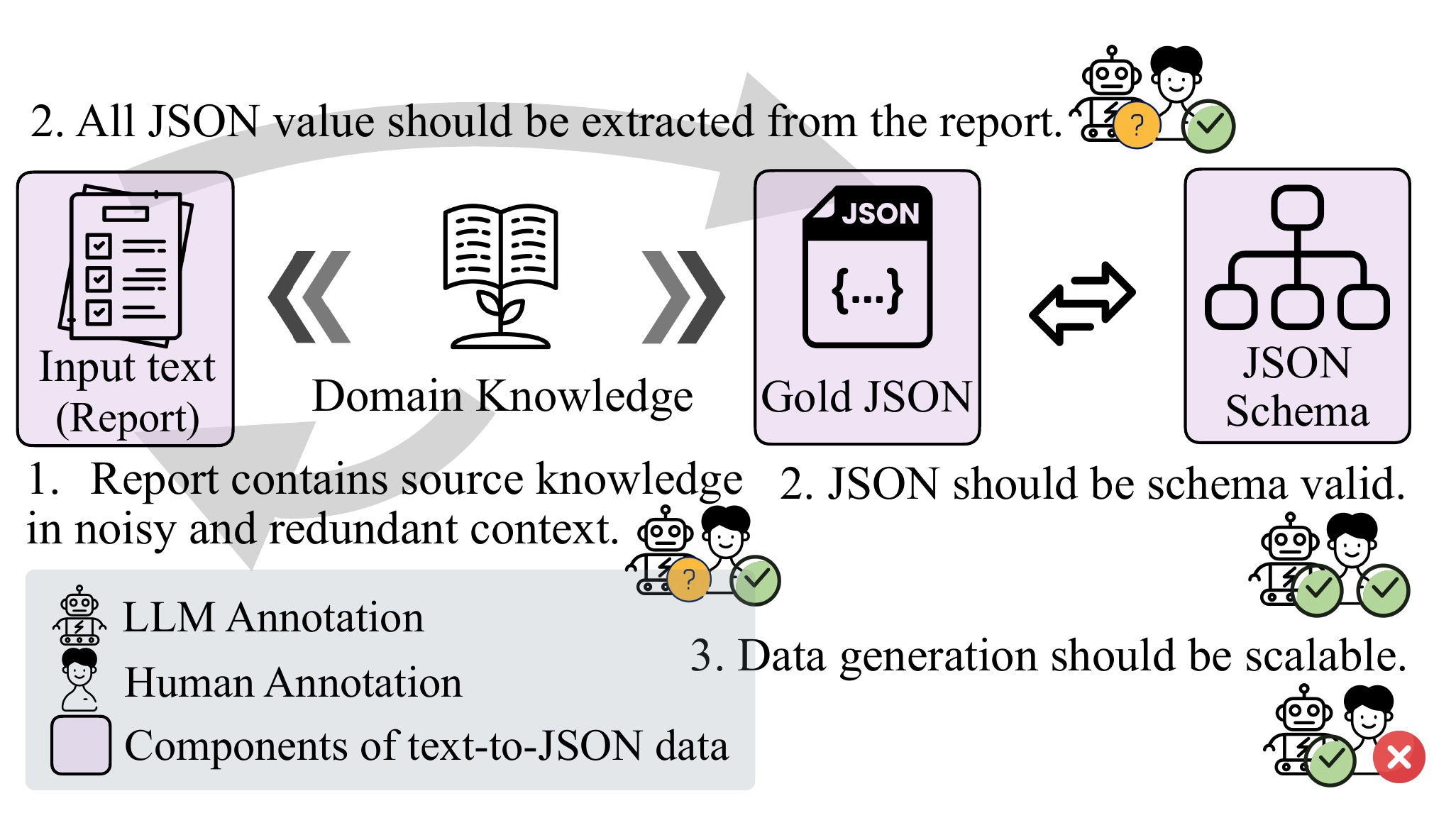}
    \caption{Desired properties and trade-offs in text-to-JSON data construction. A good instance needs a grounded report, a verifiable JSON, and a scalable pipeline, yet LLM and human annotation each satisfy only part of these, motivating a source-grounded approach.}
    \label{fig:1}
    \vspace{2.8\baselineskip}
\end{wrapfigure}

Figure~\ref{fig:1} summarizes three key requirements for high-quality text-to-JSON data: (a) each instance should contain a realistic input report, (b) each instance should pair the report with a schema-valid and verifiable JSON target, and (c) the overall construction process should be scalable.
However, existing data construction strategies face a fundamental trade-off between scalability and verifiability. While LLM-based annotation of real reports is scalable, its non-deterministic nature offers limited guarantees that the generated JSON is grounded in the source report or verifiable against reliable domain knowledge. Human annotation provides stronger quality control, but its cost and limited scalability make it difficult to construct long, complex reports that resemble real-world scenarios.

To address this issue, we propose \textbf{\method}, a scalable data generation pipeline for text-to-JSON learning.
\method{} uses spreadsheets as shared sources of domain knowledge to construct both reports and JSON/schema targets. Although LLMs are used for scalable synthesis, the flow of ground-truth knowledge remains unaffected by unconstrained LLM generation.
We also introduce \textbf{\bench{}}, a benchmark built from our pipeline, consisting of 18K training examples and 851 test examples with realistic reports, JSON schemas, and gold JSON. Finally, we validate \method{} through fine-tuning experiments that compare it against existing construction approaches and real-world scenarios.
Results on \bench{} show that \method{} provides more effective supervision than existing construction approaches: it nearly doubles Qwen3-4B exact match from 38.03 to 74.27 and raises value accuracy from 68.83 to 90.69, and yields consistent gains across multiple benchmarks and base models. 

\section{Related Work}

\paragraph{Data Construction for Structured Output Generation}~The construction of scalable training and evaluation data has been studied across structured generation tasks, including Text-to-SQL~\citep{yu2018spider} and KGQA/Text-to-SPARQL~\citep{dubey2019lc}.
To improve data quality, existing approaches ground data generation in an appropriate source of knowledge~\citep{duan2025dsqg, kosten2023spider4sparql} or apply post-hoc verification to filter LLM-generated examples~\citep{caferouglu2025sing, cattaneo2025ground}.
For text-to-JSON, SLOT synthesizes training examples with LLMs and filters them through structural validation and LLM-based checks of content fidelity~\citep{wang2025slot}.
DeepJSONEval constructs evaluation instances from aggregated web content and nested schemas, using expert annotation and human review to establish ground-truth targets~\citep{zhou2025deepjsoneval}.
We build on these principles in the text-to-JSON setting, where each training instance pairs a source report with a schema and a faithful JSON target.
Specifically, we use spreadsheets as shared sources of knowledge for constructing both reports and JSON targets.
This design allows the generated artifacts to be verified against the same underlying source, combining scalable LLM-based synthesis with source-grounded supervision.

\section{Data Generation Pipeline}
As illustrated in Figure~\ref{fig:1}, a good text-to-JSON instance requires a report containing rich domain knowledge in a noisy, redundant context, together with a schema-valid JSON target that is verifiable against the report. Moreover, the overall data generation pipeline should be scalable. In this section, we detail our data collection methodology (Section~\ref{sec:data-collection}) and dataset statistics (Section~\ref{sec:dataset-stats}).
\begin{figure*}[t]
    \centering
    \includegraphics[width=\linewidth]{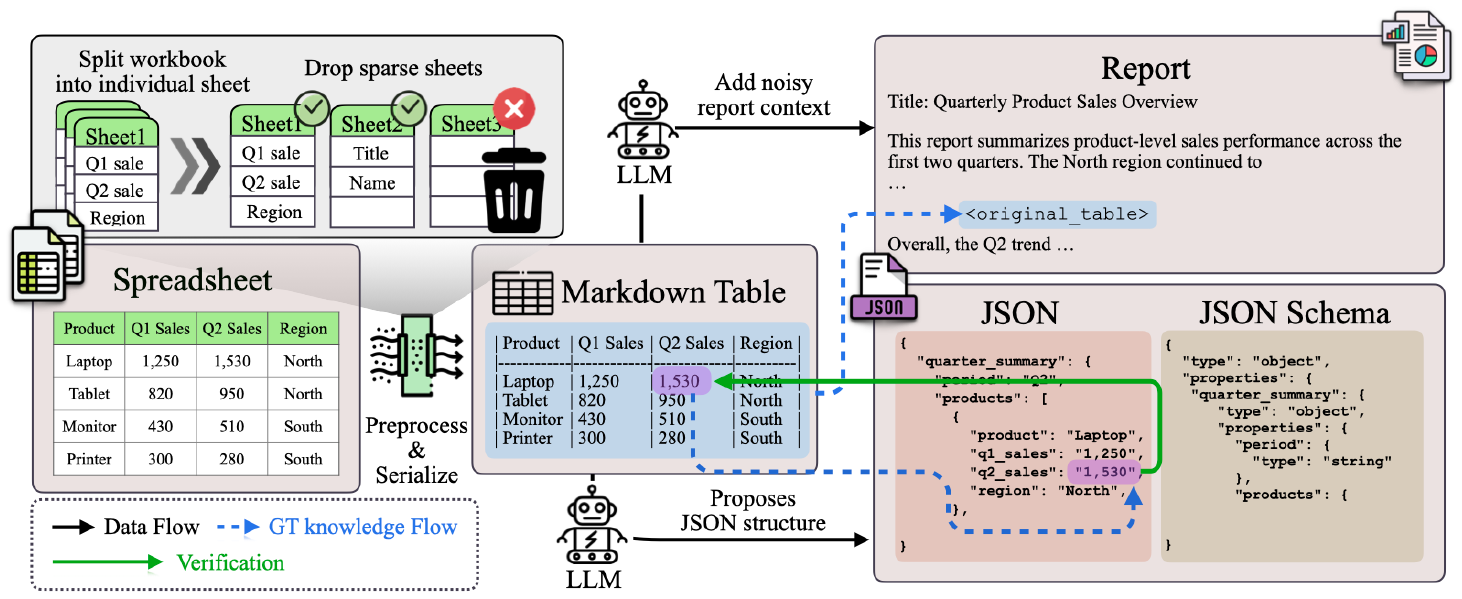}
    \caption{\textbf{Overview of \method.} A spreadsheet is filtered, serialized into a Markdown table, and used as the shared source for both report and JSON/schema generation. LLMs add surrounding context to the report and propose JSON structures. Every JSON value is verified against the source spreadsheet, ensuring that ground-truth knowledge stays anchored to the source.}
    \label{fig:2}
\end{figure*}
\subsection{Data Collection}\label{sec:data-collection}
We notice that spreadsheets are suitable for the role of source knowledge in this framework, in that they are semi-structured documents that have long served as a common medium for storing, managing, and communicating information~\citep{smith2017spreadsheet, broman2018data}. Their row--column--cell organization provides an explicit structure from which values can be extracted and analytically verified, while their widespread use makes web-scale spreadsheet corpora a rich source of domain knowledge. However, spreadsheets do not inherently provide the corresponding natural-language reports needed for text-to-JSON learning. Unlike approaches that begin with existing reports and rely on expert verification of extracted targets, our framework allows source values to be extracted and verified programmatically but requires synthesizing the reports. This shifts the data construction challenge from annotating structured targets to generating reports that faithfully convey verifiable source information.

Figure~\ref{fig:2} illustrates the data generation pipeline of \method{}. We collect spreadsheets from Sheetpedia~\citep{tian2026sheetpedia} and supplement them through keyword-based web crawling. After collection, we split each file into individual sheets and discard those with a non-empty cell ratio below 60\%, as they may lack sufficient information for report generation. Each retained sheet is serialized as a Markdown table, which serves as the shared source for both report and JSON/schema generation. We use \texttt{gemini-3-flash}~\citep{google2025gemini3flash} and \texttt{gpt-4.1}~\citep{openai2025gpt41} as generators, and the same model produces both the report and the JSON/schema target for each example. For report generation, we prompt the model to synthesize surrounding narrative and insert \texttt{<original\_table>} placeholders, which we subsequently replace with the original Markdown table to preserve the source values. For JSON/schema generation, the model proposes a JSON structure, candidate values, and a corresponding schema based on the same table. We apply deterministic verification by flattening each JSON target into leaf-level path--value pairs and checking the values against the inserted table text.
A value passes if its normalized form appears in the source text, and we discard examples with any remaining value that fails both checks.
Appendix~\ref{app:json_value_verification} provides further details.

\subsection{Dataset Statistics}\label{sec:dataset-stats}

The completed STAGE dataset contains 19,411 examples and 18,771 normalized schema specifications. The examples cover a diverse array of fields including education, public administration, healthcare, finance, transportation, and environmental data. 

\begin{table}[h]
\centering
\caption{Statistics of the complete datasets. Token lengths and primitive value counts are means, depth is the median.}
\label{tab:dataset-statistics}
\small
\begin{tabular}{@{}lrrrrrr@{}}
\toprule
Dataset & Examples & Schemas & Source tokens & JSON tokens & Primitive values & Depth \\
\midrule
\textbf{STAGE} & 19,411 & \textbf{18,771} &
2,307 & \textbf{997} & \textbf{95.0} & 3 \\
DeepJSONEval & 2,100 & 525 &
1,660 & 124 & 12.7 & 5 \\
ScrapeGraphAI & 93,695 & 5,470 &
6,161 & 173 & 19.0 & 2 \\
\bottomrule
\end{tabular}
\end{table}

As shown in Table~\ref{tab:dataset-statistics}, STAGE has the largest average JSON output size and the most primitive values per example among these releases, containing approximately $7.5\times$ and $5.0\times$ more values than DeepJSONEval and ScrapeGraphAI, respectively. Its distinguishing characteristic is therefore output breadth, while DeepJSONEval emphasizes deeper nesting and ScrapeGraphAI provides longer source contexts. Because STAGE generates both the report and JSON structure from a shared spreadsheet, these characteristics are also configurable. With sufficiently rich source tables, larger generation budgets and appropriate instructions could support longer reports, broader targets, and deeper structures, subject to the same source verification checks.

\section{Experiment Setup}
\label{sec:setup}

\paragraph{Evaluation Benchmarks.}
For Table~\ref{tab:results}, we use DeepJSONEval~\citep{zhou2025deepjsoneval} and \bench{} for evaluation. The former serves as an out-of-distribution benchmark, evaluating JSON generation under complex schemas and testing whether our approach remains effective beyond our own data distribution. The latter is a test set derived from \method{} and measures performance in the source-grounded text-to-JSON setting targeted by this work.
For Figure~\ref{fig:main_compare}, we use \bench{} to evaluate alternative strategies.
For Figure~\ref{fig:realworld}, we use two real-world benchmarks to test whether \method{}'s performance transfers beyond generated data. RealKIE-FCC-Verified~\citep{realkie2023} is a dataset built from real political-advertising invoices filed with the FCC. ExtractBench~\citep{zhang2026extractbench} is a benchmark built from real PDF documents drawn from public administrative and business sources, such as IRS tax forms, state oil-and-gas regulatory filings, vehicle valuation reports, and government procurement schedules. We use the 74 RealKIE-FCC invoices and the 237 of the 244 digital-text ExtractBench documents that fit a 131k-token context. ExtractBench is decoded with a 131k-token YaRN context so that the 32 documents whose prompt exceeds the native 32k window are included, and RealKIE with a 16,384-token generation budget because its long invoices carry hundreds of line items (Appendix~\ref{app:realworld} for details).

\paragraph{Metrics for \bench.}
For \bench, we report five metrics. \textbf{Parse Failure Rate (PFR, $\downarrow$)} measures the percentage of samples for which no valid JSON prediction can be extracted or parsed. \textbf{Schema Compliance Rate (SCR, $\uparrow$)} measures the percentage of all predictions that parse and satisfy the given JSON schema.
\textbf{Exact Match Rate (EMR, $\uparrow$)} measures the percentage of predictions whose parsed JSON object exactly matches the gold JSON after canonicalization (key sorting and whitespace normalization). \textbf{Value Accuracy (VA, $\uparrow$)} measures the fraction of gold leaf values that exactly match the prediction at the same JSON path. Finally, \textbf{Noise Ratio (NR, $\downarrow$)} measures the fraction of predicted keys that are not included in the schema, serving as an auxiliary measure of schema-irrelevant output noise. 

\begin{table*}[t]
\centering
\resizebox{\textwidth}{!}{
\begin{tabular}{l  ccc ccccc}
\toprule
 & \multicolumn{3}{c}{\textbf{DeepJSONEval}}
 & \multicolumn{5}{c}{\textbf{\bench}} \\
\cmidrule(lr){2-4} \cmidrule(lr){5-9}
\textbf{Models} & format & detailed & strict & {\shortstack{PFR$\downarrow$}}
 & \multicolumn{1}{c}{\shortstack{EMR$\uparrow$}}
 & \multicolumn{1}{c}{\shortstack{SCR$\uparrow$}}
 & \multicolumn{1}{c}{\shortstack{NR$\downarrow$}}
 & \multicolumn{1}{c}{\shortstack{VA$\uparrow$}}\\
\midrule
Qwen3-4B &
\meanstd{91.44}{0.51} & \meanstd{83.37}{0.49}  & \meanstd{51.91}{0.70}
& \textbf{\bestmeanstd{0.12}{0.00}} & \meanstd{38.03}{0.15} & \meanstd{91.15}{0.29} & \meanstd{5.61}{0.16}
  & \meanstd{68.83}{0.12}
  \\
Qwen3-4B-Thinking-2507 &
\meanstd{79.97}{0.52} & \meanstd{73.30}{0.51} & \meanstd{47.13}{0.55}
& \meanstd{70.39}{0.00} & \meanstd{14.45}{0.00} & \meanstd{19.04}{0.00} & \meanstd{77.40}{0.00}
  & \meanstd{17.13}{0.00}
  \\
\textbf{Qwen3-4B SFT  + \method} &  \textbf{\bestmeanstd{93.38}{0.00}} & \textbf{\bestmeanstd{84.68}{0.06}} & \textbf{\bestmeanstd{62.17}{0.06}}
& \meanstd{0.35}{0.00} & \textbf{\bestmeanstd{74.27}{0.00}} & \textbf{\bestmeanstd{98.24}{0.00}} & \textbf{\bestmeanstd{1.29}{0.00}}
  & \textbf{\bestmeanstd{90.69}{0.87}}
  \\
  \midrule
Qwen2.5-3B
  & \meanstd{47.52}{1.07} & \meanstd{41.31}{0.80} & \meanstd{21.54}{0.88}
& \meanstd{10.03}{1.19} & \meanstd{21.5}{0.00} & \meanstd{56.56}{0.14} & \meanstd{25.02}{0.64}
  & \meanstd{45.87}{0.01}
  \\
\textbf{Qwen2.5-3B + \method}
  & \bestmeanstd{92.68}{0.32} & \bestmeanstd{80.16}{0.14} & \bestmeanstd{41.62}{0.39}
& \bestmeanstd{0.47}{0.00} & \bestmeanstd{60.87}{0.00} & \bestmeanstd{96.36}{0.00} & \bestmeanstd{2.5}{0.00}
  & \bestmeanstd{84.95}{0.00}
  \\
\midrule

Llama-3.2-1B-Instruct
  &\meanstd{2.25}{0.40} & \meanstd{1.41}{0.22} & \meanstd{0.22}{0.08}
& \meanstd{12.46}{0.00} & \meanstd{4.07}{0.06} & \meanstd{21.58}{0.25} & \meanstd{42.18}{0.13} & \meanstd{19.54}{0.09}
  \\
\textbf{Llama-3.2-1B-Instruct + \method}
  & \bestmeanstd{38.03}{0.53} & \bestmeanstd{21.04}{0.42} & \bestmeanstd{2.76}{0.24}
& \bestmeanstd{1.18}{0.00} & \bestmeanstd{53.47}{0.00} & \bestmeanstd{93.42}{0.00} & \bestmeanstd{4.38}{0.00}
  & \bestmeanstd{80.72}{0.03}
  \\
  \midrule
Llama-3.2-3B-Instruct
 &\meanstd{22.45}{1.13} & \meanstd{18.77}{0.82} & \meanstd{9.61}{0.44}
& \meanstd{4.66}{0.06} & \meanstd{11.32}{0.06} & \meanstd{35.72}{0.00} & \meanstd{27.03}{0.11} & \meanstd{36.05}{0.06}
  \\
\textbf{Llama-3.2-3B-Instruct + \method}
  & \bestmeanstd{61.56}{0.59} & \bestmeanstd{48.34}{0.44} & \bestmeanstd{17.51}{0.02}
& \bestmeanstd{1.18}{0.00} & \bestmeanstd{59.69}{0.00} & \bestmeanstd{95.02}{0.06} & \bestmeanstd{3.53}{0.05}
  & \bestmeanstd{82.72}{0.04}
  \\

\bottomrule
\end{tabular}
}
\caption{\textbf{Evaluation results on DeepJSONEval and \bench.} For DeepJSONEval, detailed results by difficulty level are available in Appendix~\ref{app:deepjsoneval}. Scores are averaged over three independent trials, with standard deviations shown alongside. \underline{Underline} denotes the best result within each group and the \textbf{bold} entry indicates the best overall performance.}
\label{tab:results}
\end{table*}

\paragraph{Training Setup.}
For Table~\ref{tab:results}, to test whether the benefit of our data generalizes across base models, we fine-tune Qwen3-4B, Qwen2.5-3B~\citep{yang2025qwen25technicalreport}, Llama-3.2-3B-Instruct~\citep{grattafiori2024llama,meta2024llama32}, and Llama-3.2-1B-Instruct on our data, and compare each finetuned model with its non-finetuned counterpart. The \method{} training set holds 18,560 report--schema--JSON triples, and \bench{} is the disjoint test set split of 851. Every \method{} model is trained with the same full fine-tuning recipe (Table~\ref{tab:stage_sft_setting} in Appendix~\ref{app:reproducibility}).
In Figure~\ref{fig:main_compare}, we use three datasets built with alternative construction strategies to test whether our approach outperforms them.
We use 4.32K training examples from JSONSchemaBench~\citep{geng2025jsonschemabench}, and sample 20K examples each from Glaive function-calling data~\citep{glaive2024functioncallingv2} and ScrapeGraphAI-100K~\citep{brach2026scrapegraphai} as a training set.
The construction of each comparison dataset is described in Appendix~\ref{app:reproducibility}. For these three datasets, we use the same fine-tuning settings as for the \method{} model. 
\method{} fine-tuning runs use two NVIDIA H200 GPUs. Detailed reproducibility settings are specified in Appendix~\ref{app:reproducibility}.

\paragraph{Constrained Decoding.} In Figure~\ref{fig:main_compare}, we compare another structured generation strategy, constrained decoding, with our method. We decode the untrained and the \method{}-trained Qwen3-4B under grammar constraints derived from the target schema, using XGrammar~\citep{dong2025xgrammar} through vLLM. XGrammar compiles 798 of the 851 \bench{} schemas; the remaining 53 use keywords it does not support, so constrained results are reported on those 798 (Appendix~\ref{app:constrained} for more details).

\section{Results}
\subsection{Effect of Fine-Tuning on \method{}}
\label{sec:exp1}

We trained four base models to confirm whether our approach generalizes across various models. Fine-tuning on \method{} data substantially improves the model's performance, as shown in Table~\ref{tab:results}. For Qwen3-4B, exact-match accuracy improves by 36.24 points and value accuracy by 21.86 points on STAGE-Eval. Moreover, Llama-3.2-3B-Instruct gains 48.37 and 46.67 points, respectively. The gains from \method{} remain consistent across other model families, including Qwen2.5 and Llama-3.2-1B-Instruct. Additionally, the gains are consistent in the out-of-distribution setting of DeepJSONEval. Training on \method{} improves the aggregate scores for every model, with an average gain of 34.70 points for Qwen2.5-3B, 25.53 points for Llama-3.2-3B-Instruct, and 19.32 points for Llama-3.2-1B-Instruct. Even Qwen3-4B, which is already strong on this benchmark, gains 4.50 points on average, and the \method{}-trained model outperforms Qwen3-4B-Thinking-2507 on all three metrics.


\subsection{Comparison with Alternative Strategies}
\label{sec:exp2}

\begin{figure}[t]
  \centering
  \includegraphics[width=\linewidth]{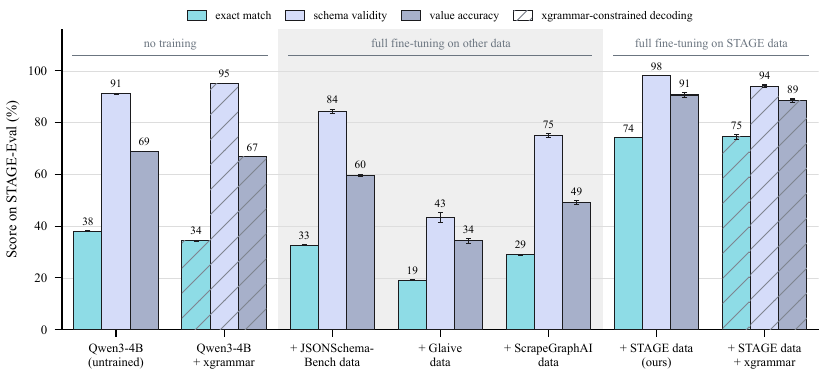}
  \caption{\textbf{\method{} data wins the data-construction strategy comparison, and training beats grammar constraints on values at no inference-time cost.} Qwen3-4B on \bench{}, temperature-0.6 sampling averaged over three seeds: untrained (left), fine-tuned on each comparison dataset with the \method{} configuration of Table~\ref{tab:stage_sft_setting} (shaded band), and fine-tuned on \method{} data (right). Hatched bars decode under XGrammar (on the 798 of 851 schemas the backend compiles), all other bars decode freely. Detailed results and inference cost in Appendix~\ref{app:constrained}.}
  \label{fig:main_compare}
\end{figure}

Additionally, we compare alternative structured output generation strategies with \method{}.
Figure~\ref{fig:main_compare} compares the same base model fine-tuned on different text-to-JSON datasets, with identical training settings. Notably, training on the three alternative datasets lowers all three evaluation metrics in \bench{}. Schema validity drops by 6.8 points with JSONSchemaBench data, by 16.2 with ScrapeGraphAI data and by 47.9 with Glaive data, and value accuracy falls by 9.2, 19.8 and 34.5 points in the same order. In contrast, \method{} data improves the model performance, 36.2 points of exact match, 7.1 of schema validity and 21.9 of value accuracy over the untrained model, suggesting that a carefully designed approach is needed to provide the model with accurate training signal.



Constrained decoding improves the model's structural consistency, improving 3.1 points in schema validity. However, value-related metrics fall slightly as a trade-off, 4.1 points in exact match and 2.2 points in value accuracy. Since a constraint can only remove irrelevant tokens, it cannot supply a value the model does not know.
On the other hand, training on \method{} data outperforms the base model by 36.2, 7.1 and 21.9 points in exact match, schema validity and value accuracy, while performing 3.0 points better on schema validity than constrained decoding.
The trained model has an advantage at inference time as well. Since constrained decoding filters every decoding step through the grammar and compiles each schema before generation, it adds about 9\% higher median latency, 15\% lower throughput and a 19 ms compile step per schema, while the trained model needs none of these steps and runs 2\% faster than the untrained model (see Table~\ref{tab:app_cost}).


\subsection{Transfer to Real-World Documents}
\label{sec:exp3}

\begin{figure}[t]
  \centering
  \begin{subfigure}[b]{0.491\linewidth}\centering
    \includegraphics[width=\linewidth]{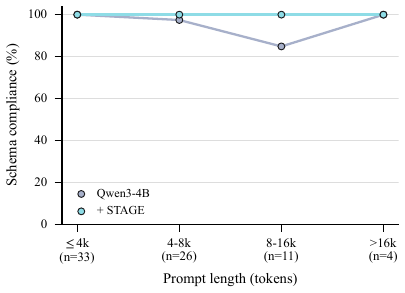}
    \caption{RealKIE-FCC invoices}\label{fig:realworld:a}
  \end{subfigure}\hfill
  \begin{subfigure}[b]{0.491\linewidth}\centering
    \includegraphics[width=\linewidth]{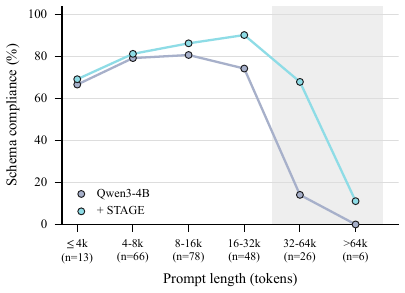}
    \caption{ExtractBench, 131k YaRN context}\label{fig:realworld:b}
  \end{subfigure}
  \caption{\textbf{On real-world documents \method{} training improves structural consistency, especially on long inputs.} Documents are grouped by prompt length (report + schema). Each point is one bucket, with bucket sizes under the axis. (a) Schema compliance of RealKIE-FCC invoices. The base model dips to 84.8\% on 8--16k-token invoices while the \method{} model stays at 100\% in every bucket.
  (b) Schema compliance in ExtractBench documents, decoded with a 131k-token YaRN context. \method{} training outperforms the base model in every bucket, and beyond the native context both models decline, but the \method{} model declines far more slowly.}
  \label{fig:realworld}
\end{figure}

Finally, we evaluate on two real-world benchmarks to test whether the \method{} model's structural consistency transfers to real-world scenarios. 
On the FCC invoices (Figure~\ref{fig:realworld}\subref{fig:realworld:a}) both models are schema-valid on short invoices, but the base model drops to 84.8\% on 8--16k-token invoices while the \method{} model stays at 100\% in every bucket.
On the ExtractBench documents (Figure~\ref{fig:realworld}\subref{fig:realworld:b}) the schema-compliance gap widens with length: 2.5 points on documents of at most 4k tokens, 2.0 on 4--8k, 5.5 on 8--16k, 16.0 on 16--32k, and 53.8 between 32k and 64k tokens. Above 64k tokens, however, where the input is already twice the 32k context the base model natively supports, both models fail on most of the six documents, but \method{} still performs better than the base model. Taken together, the advantage of \method{} training in structural consistency transfers to real-world documents, especially for longer ones. On value accuracy the picture is less clear: \method{} training helps the weaker base model, but for the stronger model the trained model is somewhat weaker on line-item values. Details are described in Appendix~\ref{app:realworld}.

\section{Conclusion}
In this paper, we present \method{}, a source-grounded data generation pipeline for text-to-JSON learning, and introduce \bench{}. Experiments show that \method{} produces stronger training data than existing construction baselines. On \bench{}, fine-tuning Qwen3-4B with \method{} nearly doubles exact match accuracy, from 38.03 to 74.27, with consistent gains on DeepJSONEval and across Qwen2.5/3 and Llama3 models. 
The STAGE model also works in real-world settings, and performs particularly well as the input document becomes longer. While this work focuses on spreadsheet-to-report-to-JSON generation, future work may extend the same source-grounded principle to broader $X$-to-JSON settings, including forms, tables, databases, knowledge graphs, and other domain-specific sources.


\bibliographystyle{plainnat}
\bibliography{anthology,custom}

@inproceedings{wang2025slot,
  title={{SLOT}: Structuring the Output of Large Language Models},
  author={Shen, Zhengyuan and Wang, Darren Yow-Bang and Mishra, Soumya Smruti and Xu, Zhichao and Teng, Yifei and Ding, Haibo},
  booktitle={Proceedings of the 2025 Conference on Empirical Methods in Natural Language Processing: Industry Track},
  pages={472--491},
  year={2025}
}

@article{yang2025qwen3,
  title={Qwen3 technical report},
  author={Yang, An and Li, Anfeng and Yang, Baosong and Zhang, Beichen and Hui, Binyuan and Zheng, Bo and Yu, Bowen and Gao, Chang and Huang, Chengen and Lv, Chenxu and others},
  journal={arXiv preprint arXiv:2505.09388},
  year={2025}
}

@misc{yang2025qwen25technicalreport,
  title        = {Qwen2.5 Technical Report},
  author       = {Yang, An and Yang, Baosong and Zhang, Beichen and Hui, Binyuan and Zheng, Bo and Yu, Bowen and Li, Chengyuan and Liu, Dayiheng and Huang, Fei and Wei, Haoran and others},
  year         = {2024},
  eprint       = {2412.15115},
  archivePrefix= {arXiv},
  primaryClass = {cs.CL},
  doi          = {10.48550/arXiv.2412.15115}
}

@article{grattafiori2024llama,
  title={The llama 3 herd of models},
  author={Grattafiori, Aaron and Dubey, Abhimanyu and Jauhri, Abhinav and Pandey, Abhinav and Kadian, Abhishek and Al-Dahle, Ahmad and Letman, Aiesha and Mathur, Akhil and Schelten, Alan and Vaughan, Alex and others},
  journal={arXiv preprint arXiv:2407.21783},
  year={2024}
}

@misc{google2025gemini3flash,
  title        = {Gemini 3 Flash: frontier intelligence built for speed},
  author       = {{Google}},
  year         = {2025},
  month        = dec,
  howpublished = {\url{https://blog.google/products-and-platforms/products/gemini/gemini-3-flash/}},
  note         = {Google Blog, Accessed: 2026-05-26}
}

@article{tian2026sheetpedia,
  title={Sheetpedia: A 300K-Spreadsheet Corpus for Spreadsheet Intelligence and LLM Fine-Tuning},
  author={Tian, Zailong and Han, Zhuoheng and Wang, Houfeng and Liao, Lizi},
  journal={Advances in Neural Information Processing Systems},
  volume={38},
  year={2025}
}

@inproceedings{zheng2024llamafactory,
  title={Llamafactory: Unified efficient fine-tuning of 100+ language models},
  author={Zheng, Yaowei and Zhang, Richong and Zhang, Junhao and Ye, Yanhan and Luo, Zheyan},
  booktitle={Proceedings of the 62nd annual meeting of the association for computational linguistics (volume 3: system demonstrations)},
  pages={400--410},
  year={2024}
}

@inproceedings{lu2022unified,
  title={Unified structure generation for universal information extraction},
  author={Lu, Yaojie and Liu, Qing and Dai, Dai and Xiao, Xinyan and Lin, Hongyu and Han, Xianpei and Sun, Le and Wu, Hua},
  booktitle={Proceedings of the 60th annual meeting of the association for computational linguistics (volume 1: long papers)},
  pages={5755--5772},
  year={2022}
}

@article{geng2025jsonschemabench,
  title={Jsonschemabench: A rigorous benchmark of structured outputs for language models},
  author={Geng, Saibo and Cooper, Hudson and Moskal, Micha{\l} and Jenkins, Samuel and Berman, Julian and Ranchin, Nathan and West, Robert and Horvitz, Eric and Nori, Harsha},
  journal={arXiv preprint arXiv:2501.10868},
  year={2025}
}

@article{zhao2023survey,
  title={A survey of large language models},
  author={Zhao, Wayne Xin and Zhou, Kun and Li, Junyi and Tang, Tianyi and Dong, Zican and Hou, Yupeng and Zhang, Beichen and Min, Yingqian and Zhang, Junjie and Liu, Peiyu and others},
  journal={Frontiers of Computer Science},
  volume={20},
  number={12},
  pages={2012627},
  year={2026},
  publisher={Springer}
}

@article{zhou2025deepjsoneval,
  title={DeepJSONEval: Benchmarking Complex Nested JSON Data Mining for Large Language Models},
  author={Zhou, Zhicheng and Li, Jing and Qiu, Suming and Huang, Junjie and Qiu, Linyuan and Sun, Zhijie},
  journal={arXiv preprint arXiv:2509.25922},
  year={2025}
}

@article{zhang2026extractbench,
  title={{ExtractBench}: A Benchmark for Schema-Guided Enterprise Document Extraction},
  author={Zhang, Boyang and Lyjak, Adrian and Stewart, Eli and Li, Zhaoqi and Suo, Simon},
  journal={arXiv preprint arXiv:2607.29677},
  year={2026}
}

@article{realkie2023,
  title={RealKIE: Five Novel Datasets for Enterprise Key Information Extraction},
  author={Townsend, Benjamin and May, Madison and Mackowiak, Katherine and Wells, Christopher},
  journal={arXiv preprint arXiv:2403.20101},
  year={2024}
}

@misc{meta2024llama32,
  title        = {Llama 3.2: Revolutionizing edge AI and vision with open, customizable models},
  author       = {{Meta AI}},
  year         = {2024},
  month        = sep,
  howpublished = {\url{https://ai.meta.com/blog/llama-3-2-connect-2024-vision-edge-mobile-devices/}}
}

@article{singh2026structured,
  title={The Structured Output Benchmark: A Multi-Source Benchmark for Evaluating Structured Output Quality in Large Language Models},
  author={Singh, Abhinav Kumar and Khurdula, Harsha Vardhan and Khemlani, Yoeven D and Agarwal, Vineet},
  journal={arXiv preprint arXiv:2604.25359},
  year={2026}
}

@article{willard2023efficient,
  title={Efficient guided generation for large language models},
  author={Willard, Brandon T and Louf, R{\'e}mi},
  journal={arXiv preprint arXiv:2307.09702},
  year={2023}
}

@inproceedings{lu2025learning,
  title={Learning to generate structured output with schema reinforcement learning},
  author={Lu, Yaxi and Li, Haolun and Cong, Xin and Zhang, Zhong and Wu, Yesai and Lin, Yankai and Liu, Zhiyuan and Liu, Fangming and Sun, Maosong},
  booktitle={Proceedings of the 63rd Annual Meeting of the Association for Computational Linguistics (Volume 1: Long Papers)},
  pages={4905--4918},
  year={2025}
}

@inproceedings{scholak2021picard,
  title={PICARD: Parsing incrementally for constrained auto-regressive decoding from language models},
  author={Scholak, Torsten and Schucher, Nathan and Bahdanau, Dzmitry},
  booktitle={Proceedings of the 2021 conference on empirical methods in natural language processing},
  pages={9895--9901},
  year={2021}
}

@article{dong2025xgrammar,
  title={Xgrammar: Flexible and efficient structured generation engine for large language models},
  author={Dong, Yixin and Ruan, Charlie F and Cai, Yaxing and Xu, Ziyi and Zhao, Yilong and Lai, Ruihang and Chen, Tianqi},
  journal={Proceedings of Machine Learning and Systems},
  volume={7},
  year={2025}
}

@article{patil2024gorilla,
  title={Gorilla: Large language model connected with massive apis},
  author={Patil, Shishir G and Zhang, Tianjun and Wang, Xin and Gonzalez, Joseph E},
  journal={Advances in Neural Information Processing Systems},
  volume={37},
  pages={126544--126565},
  year={2024}
}

@inproceedings{qin2024toolllm,
  title={Toolllm: Facilitating large language models to master 16000+ real-world apis},
  author={Qin, Yujia and Liang, Shihao and Ye, Yining and Zhu, Kunlun and Yan, Lan and Lu, Yaxi and Lin, Yankai and Cong, Xin and Tang, Xiangru and Qian, Bill and others},
  booktitle={The Twelfth International Conference on Learning Representations},
  year={2024},
  url={https://openreview.net/forum?id=dHng2O0Jjr}
}

@inproceedings{smith2017spreadsheet,
  title={Spreadsheet practices and challenges in a large multinational conglomerate},
  author={Smith, Justin and Middleton, Justin A and Kraft, Nicholas A},
  booktitle={2017 IEEE Symposium on Visual Languages and Human-Centric Computing (VL/HCC)},
  pages={155--163},
  year={2017},
  organization={IEEE}
}

@article{broman2018data,
  title={Data organization in spreadsheets},
  author={Broman, Karl W and Woo, Kara H},
  journal={The American Statistician},
  volume={72},
  number={1},
  pages={2--10},
  year={2018},
  publisher={Taylor \& Francis}
}

@inproceedings{duan2025dsqg,
  title={DSQG-syn: Synthesizing high-quality data for text-to-SQL parsing by domain specific question generation},
  author={Duan, Shaoming and Wu, Youxuan and Liu, Chuanyi and Zhang, Yuhao and Wang, Zirui and Han, Peiyi and Yu, Shengyuan and Yan, Liang and Liang, Yingwei},
  booktitle={Findings of the Association for Computational Linguistics: NAACL 2025},
  pages={2971--2989},
  year={2025}
}

@article{caferouglu2025sing,
  title={Sing-sql: A synthetic data generation framework for in-domain text-to-sql translation},
  author={Cafero{\u{g}}lu, Hasan Alp and {\c{C}}elik, Mehmet Serhat and Ulusoy, {\"O}zg{\"u}r},
  journal={arXiv preprint arXiv:2509.25672},
  year={2025}
}

@article{cattaneo2025ground,
  title={Ground-Truth Subgraphs for Better Training and Evaluation of Knowledge Graph Augmented LLMs},
  author={Cattaneo, Alberto and Luschi, Carlo and Justus, Daniel},
  journal={arXiv preprint arXiv:2511.04473},
  year={2025}
}

@inproceedings{kosten2023spider4sparql,
  title={Spider4SPARQL: a complex benchmark for evaluating knowledge graph question answering systems},
  author={Kosten, Catherine and Cudr{\'e}-Mauroux, Philippe and Stockinger, Kurt},
  booktitle={2023 IEEE international conference on big data (BigData)},
  pages={5272--5281},
  year={2023},
  organization={IEEE}
}

@misc{glaive2024functioncallingv2,
  title        = {Glaive Function Calling v2},
  author       = {{Glaive AI}},
  year         = {2023},
  howpublished = {\url{https://huggingface.co/datasets/glaiveai/glaive-function-calling-v2}},
  note         = {Hugging Face dataset}
}

@article{brach2026scrapegraphai,
  title={ScrapeGraphAI-100k: Dataset for Schema-Constrained LLM Generation},
  author={Brach, William and Zuppichini, Francesco and Vinciguerra, Marco and Padoan, Lorenzo},
  journal={arXiv preprint arXiv:2602.15189},
  year={2026}
}

@inproceedings{yu2018spider,
  title={Spider: A large-scale human-labeled dataset for complex and cross-domain semantic parsing and text-to-sql task},
  author={Yu, Tao and Zhang, Rui and Yang, Kai and Yasunaga, Michihiro and Wang, Dongxu and Li, Zifan and Ma, James and Li, Irene and Yao, Qingning and Roman, Shanelle and others},
  booktitle={Proceedings of the 2018 conference on empirical methods in natural language processing},
  pages={3911--3921},
  year={2018}
}

@inproceedings{dubey2019lc,
  title={Lc-quad 2.0: A large dataset for complex question answering over wikidata and dbpedia},
  author={Dubey, Mohnish and Banerjee, Debayan and Abdelkawi, Abdelrahman and Lehmann, Jens},
  booktitle={International semantic web conference},
  pages={69--78},
  year={2019},
  organization={Springer}
}

@misc{openai2025gpt41,
  title        = {Introducing GPT-4.1 in the API},
  author       = {{OpenAI}},
  year         = {2025},
  month        = apr,
  howpublished = {\url{https://openai.com/index/gpt-4-1/}}
}

\clearpage
\appendix

\section{Difficulty-Specific DeepJSONEval Results}
\label{app:deepjsoneval}

\begin{table*}[h]
\centering

\resizebox{\textwidth}{!}{
\begin{tabular}{l  ccc ccc}
\toprule
 & \multicolumn{6}{c}{\textbf{DeepJSONEval}} \\
\cmidrule(lr){2-7}
 & \multicolumn{3}{c}{Medium}
 & \multicolumn{3}{c}{Hard} \\
\cmidrule(lr){2-4} \cmidrule(lr){5-7}
\textbf{Models}
& format & detailed & strict
& format & detailed & strict \\
\midrule

Qwen3-4B
& \meanstd{92.22}{0.67}
& \meanstd{86.97}{0.65}
& \meanstd{60.82}{0.77}
& \textbf{\bestmeanstd{91.09}{0.52}}
& \textbf{\bestmeanstd{81.73}{0.53}}
& \textbf{\bestmeanstd{47.85}{0.68}} \\

Qwen3-4B-Thinking-2507
& \meanstd{85.27}{0.77}
& \meanstd{80.17}{0.99}
& \meanstd{56.45}{2.07}
& \meanstd{77.56}{0.66}
& \meanstd{70.18}{0.59}
& \meanstd{42.89}{0.15} \\


\textbf{Qwen3-4B SFT + \method}
& \textbf{\bestmeanstd{99.14}{0.09}}
& \textbf{\bestmeanstd{92.56}{0.03}}
& \textbf{\bestmeanstd{62.35}{0.26}}
& \meanstd{90.77}{0.04}
& \meanstd{81.11}{0.09}
& \meanstd{47.56}{0.04} \\

\midrule

Qwen2.5-3B
& \meanstd{53.41}{3.35}
& \meanstd{46.82}{2.64}
& \meanstd{23.73}{1.67}
& \meanstd{44.86}{0.04}
& \meanstd{38.81}{0.05}
& \meanstd{20.54}{0.52} \\

\textbf{Qwen2.5-3B + \method}
& \bestmeanstd{97.00}{0.62}
& \bestmeanstd{86.18}{0.23}
& \bestmeanstd{48.99}{0.64}
& \bestmeanstd{90.72}{0.19}
& \bestmeanstd{77.43}{0.21}
& \bestmeanstd{38.27}{0.28} \\

\midrule

Llama-3.2-1B-Instruct
& \meanstd{3.96}{1.31}
& \meanstd{2.59}{0.78}
& \meanstd{0.51}{0.18}
& \meanstd{1.48}{0.35}
& \meanstd{0.88}{0.18}
& \meanstd{0.09}{0.04} \\

\textbf{Llama-3.2-1B-Instruct + \method}
& \bestmeanstd{62.04}{0.66}
& \bestmeanstd{33.44}{0.94}
& \bestmeanstd{2.18}{0.09}
& \bestmeanstd{27.12}{1.06}
& \bestmeanstd{15.40}{1.04}
& \bestmeanstd{3.02}{0.40} \\

\midrule

Llama-3.2-3B-Instruct
& \meanstd{29.37}{0.88}
& \meanstd{24.06}{0.29}
& \meanstd{11.64}{0.18}
& \meanstd{19.30}{1.24}
& \meanstd{16.36}{1.06}
& \meanstd{8.68}{0.56} \\

\textbf{Llama-3.2-3B-Instruct + \method}
& \bestmeanstd{71.24}{0.32}
& \bestmeanstd{57.12}{0.29}
& \bestmeanstd{21.95}{0.26}
& \bestmeanstd{57.16}{1.00}
& \bestmeanstd{44.35}{0.78}
& \bestmeanstd{15.49}{0.08} \\

\bottomrule
\end{tabular}
}
\caption{Evaluation results on DeepJSONEval Medium and Hard splits. Scores are averaged over three runs, with standard deviations shown alongside each mean.
\underline{Underline} denotes the best result within each group and the \textbf{bold} entry indicates the best overall performance.}
\label{tab:deepjsoneval_medium_hard}
\end{table*}

Table~\ref{tab:deepjsoneval_medium_hard} shows the difficulty-specific DeepJSONEval results (medium and hard splits) for every base model and its \method{}-trained counterpart. On the hard split, the untrained Qwen3-4B scores marginally above its \method{}-trained counterpart (81.73 vs.\ 81.11 detailed, 47.85 vs.\ 47.56 strict). The differences are 0.3--0.6 points, comparable to the run-to-run variation of the untrained model (standard deviations 0.5--0.7), so the two should be read as equivalent rather than as a regression. The pattern is consistent with the main table: Qwen3-4B is already a strong JSON generator on this benchmark, so there is little headroom for training data to add, and on the medium split the gain is correspondingly modest (86.97 to 92.56 detailed). Where the base model is weak the same data makes a large difference: on the hard split, detailed scores rise from 38.81 to 77.43 for Qwen2.5-3B, from 16.36 to 44.35 for Llama-3.2-3B and from 0.88 to 15.40 for Llama-3.2-1B.

\section{Details of JSON Value Verification}
\label{app:json_value_verification}

We apply an additional deterministic verification step to ensure that generated JSON values are supported by the source table included in the report. This step is implemented as a lexical coverage check between JSON leaf values and the Markdown table text in the generated report.

First, we flatten each generated JSON object into path--value pairs. For example, a JSON object such as
\begin{verbatim}
{
  "company": "Apple",
  "items": [{"price": "1000"}]
}
\end{verbatim}
is converted into leaf-level entries such as
\begin{verbatim}
$.company = Apple
$.items[0].price = 1000
\end{verbatim}
This allows us to verify values inside nested objects and arrays, rather than checking only top-level fields.

Second, we extract the Markdown table portion from the input report. Since the original source table is inserted into the report during report generation, we restrict the verification target to lines that belong to the Markdown table, identified by the presence of the table delimiter ``|''. When a sheet heading such as \texttt{Sheet:} or \texttt{\#\# Sheet:} immediately precedes the table, we include it as part of the verification text, because the sheet name may also provide valid source information.

Third, both JSON values and table text are normalized before matching. We keep only numeric and English alphanumeric tokens and remove other symbols, punctuation, and formatting marks. The normalized strings are lowercased and joined without whitespace. For example, a value such as ``\texttt{\$1,000}'' is normalized to ``1000'', and ``Apple Inc.'' to ``appleinc''. This normalization makes the check robust to superficial formatting differences such as punctuation, commas, spaces, and case.

The matching procedure has two stages. We first perform substring matching: a JSON value passes verification if its normalized form appears in the normalized text of the Markdown table. This captures cases where the value appears directly in the table, possibly with minor formatting differences. If substring matching fails, we apply a token-level fallback. The JSON value is split into normalized tokens, and each token is checked against the set of tokens appearing in the table. The value is considered covered if all of its tokens appear somewhere in the table. For instance, a value such as ``Apple Inc'' can pass the fallback check if the tokens ``apple'' and ``inc'' both appear in the table text.

We exclude \texttt{null}, boolean values, and empty strings from this coverage check, since such values are often not literally expressed in the table and are ambiguous as evidence-bearing values. For all other leaf values, the verifier records whether the value is covered by the table. If any checked leaf value is not covered, the example is discarded from the final dataset.

This procedure is intentionally lexical rather than semantic. It is designed as a conservative and scalable filter for checking whether each JSON value has explicit textual evidence in the inserted Markdown table. It does not attempt to verify paraphrases, unit conversions, arithmetic transformations, date normalization, or other forms of semantic equivalence. Therefore, the verifier may reject some semantically valid transformed values, but retained examples have a strong guarantee that their JSON leaf values are explicitly supported by the table text.

\section{Reproducibility Details}
\label{app:reproducibility}

This section provides the details needed to reproduce our fine-tuning and evaluation experiments. Fine-tuning uses random seed 42; every evaluation is run with seeds 42, 43 and 44 and reported as the mean over the three runs. Fine-tuning experiments are conducted on two NVIDIA H200 GPUs, and inference is performed on a single NVIDIA H200 GPU. 

\paragraph{Base models.}
We evaluate the following open-weight base models: \texttt{Qwen/Qwen3-4B}~\citep{yang2025qwen3}, \texttt{Qwen/Qwen2.5-3B}, \texttt{meta-llama/Llama-3.2-1B-Instruct}, and \texttt{meta-llama/Llama-3.2-3B-Instruct}; the reasoning-model reference row of Table~\ref{tab:results} is \texttt{Qwen/Qwen3-4B-Thinking-2507}, evaluated without fine-tuning. We use the corresponding Hugging Face tokenizers and chat templates for each model. For Qwen3-4B, we use the \texttt{qwen3\_nothink} chat template to disable thinking-mode outputs for JSON generation, for the untrained model as well as for the trained ones.

\paragraph{Fine-tuning setting for \method{}.}
For \method{}, we use standard supervised fine-tuning with full-parameter updates. We use LlamaFactory~\citep{zheng2024llamafactory} for training. The maximum sequence length is set to 8192 tokens. Table~\ref{tab:stage_sft_setting} summarizes the configuration.

\begin{table}[h]
\centering
\small
\begin{tabular}{ll}
\toprule
Setting & Value \\
\midrule
Training framework & LLaMA-Factory \\
Stage & SFT \\
Fine-tuning type & Full fine-tuning \\
Base model example & \texttt{Qwen/Qwen3-4B} \\
Chat template & \texttt{qwen3\_nothink} \\
Cutoff length & 8192 \\
Max samples & 50,000 \\
Validation split & 5\% \\
Epochs & 3 \\
Per-device train batch size & 4 \\
Gradient accumulation steps & 8 \\
Learning rate & $4.0\times10^{-5}$ \\
LR scheduler & Cosine \\
Warmup ratio & 0.1 \\
Optimizer & \texttt{adamw\_torch\_fused} \\
Precision & bf16 \\
Random seed & 42 \\
Preprocessing workers & 16 \\
Dataloader workers & 4 \\
Flash attention setting & \texttt{sdpa} \\
Liger kernel & Disabled \\
Gradient checkpointing & Enabled \\
DeepSpeed & Not used \\
\bottomrule
\end{tabular}
\caption{Fine-tuning configuration for \method{} training.}
\label{tab:stage_sft_setting}
\end{table}

\paragraph{Construction of the comparison datasets.}
We construct a schema-first synthetic baseline from JSONSchemaBench~\citep{geng2025jsonschemabench}, which provides JSON schemas without reports or gold JSON targets. For each schema, we synthesize a schema-valid JSON object and a short report using the Qwen3-4B model: the model first fills the schema with a gold JSON, then writes a report that contains every gold value; we keep the 4,324 pairs (of 6,000 attempts over the 5.75K training schemas) whose JSON validates against the schema and whose values are all found in the report, the same verification we apply to \method{} data. For Glaive function-calling data~\citep{glaive2024functioncallingv2}, we treat function call arguments as gold JSON and use the remaining dialogue context, including the function definition corresponding to the JSON schema, as the input report. ScrapeGraphAI-100K~\citep{brach2026scrapegraphai} is already provided in a text-to-JSON format, with webpage Markdown, prompts, JSON schemas, and LLM-generated structured responses, so we use it directly as a training dataset, truncating pages that exceed the training cutoff. All three are cast into the same report--schema--JSON prompt format as \method{}, and 20,000 examples are sampled from Glaive and from ScrapeGraphAI.

\paragraph{Fine-tuning setting for baseline datasets.}
All comparison datasets (JSONSchemaBench, Glaive, ScrapeGraphAI) are trained with the hyperparameters of Table~\ref{tab:stage_sft_setting}: full-parameter fine-tuning of Qwen3-4B with the thinking-disabled chat template, an 8192-token cutoff, three epochs, learning rate $4\times10^{-5}$ with cosine decay and 0.1 warmup, bf16, and a 5\% validation split.

\paragraph{Inference setting.}
All inference experiments are performed on a single NVIDIA H200 GPU. Unless otherwise specified, inference uses vLLM with temperature $0.6$, top-$p$ $1.0$, a maximum generation length of 3100 tokens, and a maximum model length of 8192 tokens; the real-world runs raise the model length to 40,960 (RealKIE-FCC) and 131,072 with YaRN (ExtractBench), and RealKIE-FCC uses a 16,384-token generation length. Table~\ref{tab:inference_setting} summarizes the inference configuration.

\begin{table}[h]
\centering
\small
\begin{tabular}{ll}
\toprule
Setting & Value \\
\midrule
Inference engine & vLLM \\
GPU & 1 NVIDIA H200 \\
Temperature & 0.6 \\
Top-$p$ & 1.0 \\
Max new tokens & 3100 (16,384 for RealKIE-FCC) \\
Max model length & 8192 (40,960 RealKIE-FCC; 131,072 ExtractBench) \\
Random seed & 42, 43, 44 \\
\bottomrule
\end{tabular}
\caption{Inference setting used for evaluation.}
\label{tab:inference_setting}
\end{table}

\section{Training versus Grammar-Constrained Decoding}
\label{app:constrained}
\begin{table}[h]
\centering
\footnotesize
\setlength{\tabcolsep}{4pt}
\begin{tabular}{llccccc}
\toprule
\textbf{Model} & \textbf{Decoding} & PFR$\downarrow$ & EMR$\uparrow$ & SCR$\uparrow$ & NR$\downarrow$ & VA$\uparrow$ \\
\midrule
Qwen3-4B & free & 0.1 & 38.4 & 92.1 & 4.7 & 69.1 \\
Qwen3-4B & xgrammar & \textbf{0.2} & 34.3 & 95.2 & \textbf{0.6} & 66.9 \\
\midrule
Qwen3-4B + \method & free & 0.3 & \textbf{74.3} & \textbf{98.2} & 1.3 & \textbf{90.7} \\
Qwen3-4B + \method & xgrammar & 1.1 & 74.5 & 94.1 & 1.7 & 88.5 \\
\bottomrule
\end{tabular}
\caption{\textbf{Grammar-constrained decoding raises structure but not values; \method{} training raises both.} \bench{}, temperature-0.6 sampling, thinking disabled in every row. The untrained rows are scored on the 798 of 851 test schemas that xgrammar compiled in that run (seeds 42, 43 and 44); the \method{} free-decoding row is the Table~\ref{tab:results} run and the \method{} xgrammar row the same checkpoint decoded under the grammar (seeds 42, 43 and 44), both on all 851 examples. PFR is the parse-failure rate.}
\label{tab:app_constrained}
\end{table}

A common alternative to training is to constrain decoding with the target JSON Schema at inference time. We compare the two on \bench{} with vLLM's xgrammar backend, restricted to the 798 of 851 test schemas that xgrammar compiles; the remaining 53 use \texttt{format}, \texttt{multipleOf}, or \texttt{uniqueItems} keywords that the backend does not support. Table~\ref{tab:app_constrained} reads the same way for structure and values (temperature-0.6 sampling averaged over three runs, as in Table~\ref{tab:results}): with thinking disabled the untrained Qwen3-4B already parses almost every output, and the grammar raises schema validity by three points while lowering exact match by four points and value accuracy by two. \method{} training with free decoding raises exact match by 36 points and value accuracy by 22 points over the same base (Table~\ref{tab:results}), and adding the grammar on top of the trained model lowers schema validity by four points and value accuracy by two while leaving exact match unchanged, because the grammar can only remove tokens the trained model would have emitted correctly.
 
\begin{figure}[h]
  \centering
  \includegraphics[width=\linewidth]{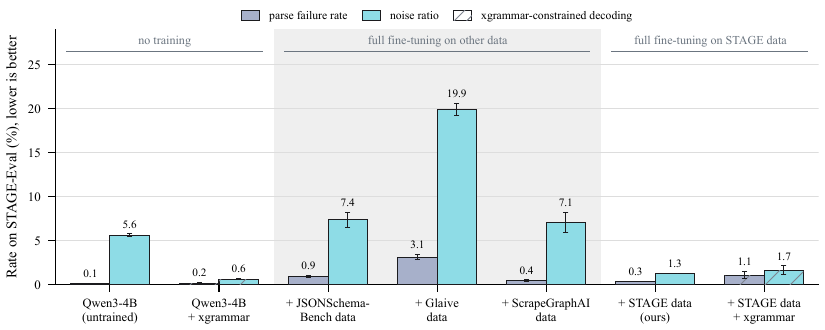}
  \caption{\textbf{The two remaining \bench{} metrics for the conditions of Figure~\ref{fig:main_compare}.} Parse failure rate and noise ratio (share of predicted keys outside the schema), lower is better; same protocol, grouping, hatching and error bars as Figure~\ref{fig:main_compare}. With thinking disabled, parse failures are rare in every condition (at most 3.1\% for the Glaive-trained model); noise mostly separates the arms: grammar constraints and \method{} training both remove almost all schema-irrelevant keys (0.6--1.7\%), while the Glaive-trained model emits them in a fifth of its keys.}
  \label{fig:app_pfr_nr}
\end{figure}
Figure~\ref{fig:app_pfr_nr} adds the two remaining \bench{} metrics for every condition of Figure~\ref{fig:main_compare}. Parse failures are rare throughout (at most 3.1\% for the Glaive-trained model), so the conditions separate on noise: grammar constraints and \method{} training both remove almost all schema-irrelevant keys (0.6--1.7\% of predicted keys), while the Glaive-trained model emits them in a fifth of its keys.

\begin{table}[h]
\centering
\footnotesize
\setlength{\tabcolsep}{3pt}
\begin{tabular}{llccccc}
\toprule
\textbf{Model} & \textbf{Decoding} & \shortstack{Latency\\median (s)} & \shortstack{Latency\\p90 (s)} & \shortstack{Throughput\\(ex/s, batch 32)} & \shortstack{Generated\\tokens} & \shortstack{Grammar\\compile (ms)} \\
\midrule
Qwen3-4B & free & 1.82 & 6.05 & 2.52 & 588 & -- \\
Qwen3-4B & xgrammar & 1.98 & 6.31 & 2.15 & 622 & 19.2 \\
\midrule
Qwen3-4B + \method & free & 1.78 & 5.61 & 2.66 & 560 & -- \\
Qwen3-4B + \method & xgrammar & 1.89 & 5.82 & 2.34 & 593 & 19.8 \\
\bottomrule
\end{tabular}
\caption{\textbf{\method{} training adds no inference-time cost; grammar constraints add about 9\% latency and cost about 15\% throughput.} Same 798 examples and one H200 as Table~\ref{tab:app_constrained}; batch-1 latency is the warm-cache pass, grammar compile time is measured per schema with the xgrammar cache disabled. Per-token time is about 4.5\,ms in every row, so latency differences follow generated length.}
\label{tab:app_cost}
\end{table}

Table~\ref{tab:app_cost} gives the cost side. Per-token time is about 4.5\,ms in every row, the trained model generates slightly fewer tokens than the base, and for the untrained model the grammar adds a per-schema compile step, about nine percent more batch-1 latency, and fifteen percent less throughput (seven and twelve percent for the trained model) on top of the 6.2\% of schemas it cannot express. The trained model therefore reaches the higher structure and value accuracy at no inference-time cost.

\section{Real-World Documents}
\label{app:realworld}\label{app:extractbench}\label{app:realkie}
\begin{figure}[h]
  \centering
  \includegraphics[width=\linewidth]{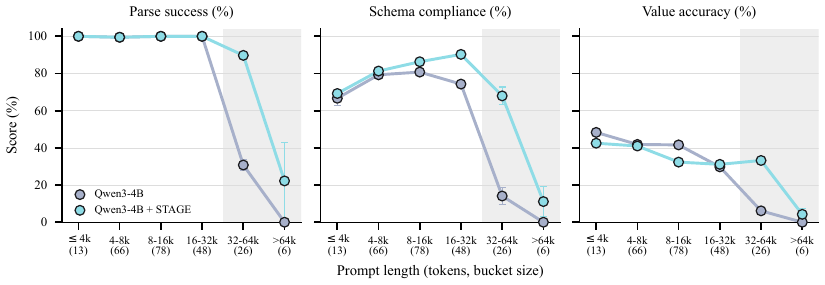}
  \caption{\textbf{ExtractBench with a 131k-token YaRN context.} All 237 digital-text documents, temperature-0.6 sampling with a 3,100-token generation budget; points are means over three seeds with standard-deviation bars, bucket sizes in parentheses, shaded region beyond the native 32k context. \method{} training raises schema compliance in every bucket and keeps most outputs schema-valid between 32k and 64k tokens, where the untrained model breaks down; value accuracy is a tie or a small deficit within the native context and is dominated by the size of the target schemas beyond it.}
  \label{fig:app_eb131}
\end{figure}
\paragraph{ExtractBench.} ExtractBench~\citep{zhang2026extractbench} pairs PDF documents with a schema and gold JSON. Figure~\ref{fig:app_eb131} scores all 237 digital-text documents by prompt length with the 131k-token context of Figure~\ref{fig:realworld}\subref{fig:realworld:b}; the numbers in this paragraph are for the subset within the native context. We take the 244 documents whose PDF text is digital, keep the 200 whose prompt fits a 32,768-token context together with the 3,100-token generation budget (205 fit the context alone; five more are excluded once the generation budget is reserved), and drop six schemas that xgrammar cannot compile, so every condition below scores the same 194 documents. Outputs are parsed by taking the first balanced JSON object in the response; with this parser every Qwen3-4B condition parses 99.5--100\% of its outputs, so the differences are in schema compliance and values. The numbers below are the temperature-0.6 runs averaged over three seeds; a greedy re-run of the same conditions gives 80.9\% schema validity and 41.2 value accuracy untrained against 82.5\% and 35.5 trained, so under greedy decoding the schema-validity gain shrinks to under two points and the value-accuracy deficit remains. \method{} training raises schema compliance from 78.0\% to 84.9\% for Qwen3-4B (67.8\% to 81.3\% on the medium split, whose documents exceed the 8,192-token training window). Value accuracy moves the other way for Qwen3-4B: the untrained model is 4.6 points higher overall (39.1 vs.\ 34.5), the gap coming entirely from the short split (41.4 vs.\ 34.9) with a tie on the medium split (33.4 vs.\ 33.6).

\begin{figure}[h]
  \centering
  \includegraphics[width=\linewidth]{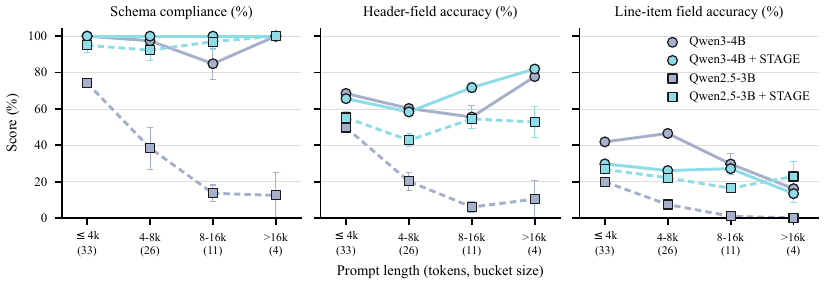}
  \caption{\textbf{RealKIE-FCC by prompt length, for two base models.} 74 invoices, temperature-0.6 sampling with the 16,384-token generation budget of Figure~\ref{fig:realworld}\subref{fig:realworld:a}; points are means over three seeds with standard-deviation bars, bucket sizes in parentheses. Circles and solid lines are Qwen3-4B, squares and dashed lines Qwen2.5-3B. For Qwen3-4B the two models tie on header fields up to 8k tokens and the trained model leads beyond; it returns a schema-valid object in every bucket, while the untrained model recovers more line items. For the smaller Qwen2.5-3B the untrained model collapses beyond 4k tokens on every metric, and \method{} training restores schema compliance to 92--100\% and lifts header and line-item accuracy in every bucket.}
  \label{fig:app_rk_budget}
\end{figure}
\paragraph{RealKIE-FCC invoices.} RealKIE-FCC-Verified~\citep{realkie2023} pairs 75 multi-page political-advertising invoices from the FCC public file with OCR text, a shared JSON Schema, and verified gold JSON; the schema has six header fields and a \texttt{LineItems} array of five-field items (median 16 items per invoice, up to 392), which makes it the most deeply nested target we evaluate. We keep the 74 documents that fit a 40,960-token context and map the schema's non-standard \texttt{float} type to \texttt{number} so that the validator accepts it. Because a leaf-level value accuracy on this schema is dominated by line items and is order-sensitive, we report header-field accuracy and line-item field accuracy after matching predicted items to gold items by field overlap. 

Figure~\ref{fig:app_rk_budget} shows both by prompt length under the 16,384-token budget, averaged over three seeds. On schema compliance and header-field accuracy the picture is consistent for both base models: the gain from \method{} training grows with document length, and for the smaller Qwen2.5-3B it is large in every bucket (schema compliance 49.3\% to 94.6\% and header-field accuracy 30.7 to 50.6 overall, with the untrained model falling below 20 on every metric beyond 8k tokens). On line-item field accuracy, the value-level metric, the claim does not hold as clearly. For Qwen3-4B the untrained model is ahead on short invoices and the two only converge as invoices grow long, and for Qwen2.5-3B the trained model is ahead throughout; above all, no model in either family exceeds 50 on this metric, so the absolute accuracy itself is what is lacking. 

We see two reasons, both properties of the current training data rather than of the method. First, \method{} reports are short (1--4k tokens) and contain few repeated records, so the trained model has never had to produce a long output and closes its item list early on invoices with dozens of items instead of emitting every value. Second, every value in \method{} training targets is copied from a verified table cell and no target field is left empty, so the model was never trained to output null or empty values where the document has none. \textbf{Both are technical limitations of the data rather than of the pipeline}: longer source tables yield longer reports and targets with more records under the same verification, and null-valued fields can be added to the targets by the same procedure, which we leave to future work.

\section{Limitations}
\label{app:limitations}
This work has several limitations. First, our experiments are conducted mainly on relatively small open-weight models. This is due to the computational cost of fine-tuning on long report-style inputs with JSON/schema targets, whose length can reach up to 8192 tokens, as well as the need to run repeated experiments across multiple baselines and benchmarks. 
Nevertheless, we believe the improvements on smaller models are meaningful, as on-device models and agents depend on achieving practically usable performance under cost and latency constraints. Further, since these models have limited text-to-JSON capability, the effect of training data is more directly observable.
Future work should further examine the scaling behavior of \method{} on larger models and broader model families.
Second, \method{} uses spreadsheets as a source of knowledge, but not all real-world domain knowledge is stored in spreadsheet form. Thus, our current pipeline is best suited to settings where tabular knowledge is explicitly available, and does not fully cover knowledge contained only in free-form text, PDFs, databases, knowledge graphs, or multimodal documents. Our main contribution, however, is not limited to spreadsheets themselves, but lies in the broader source-grounded construction principle: generating reports and JSON/schema targets from a shared verifiable source. This principle can be extended to other structured or semi-structured sources.

\section{Dataset Release}
\label{app:release}
We release \bench{} on Hugging Face (\url{https://huggingface.co/datasets/boradorish/STAGE-eval}) and its evaluation code on GitHub (\url{https://github.com/boradorish/STAGE-Eval}) under the MIT License.

\end{document}